\documentclass{bmvc2k}

\usepackage{multirow}
\usepackage{multicol}
\usepackage{algorithm,algpseudocode}

\title{Instance Segmentation of Dense and Overlapping Objects via Layering}

\addauthor{Long Chen}{long.chen@lfb.rwth-aachen.de}{1}
\addauthor{Yuli Wu}{yuli.wu@lfb.rwth-aachen.de}{1}
\addauthor{Dorit Merhof}{dorit.merhof@ur.de}{2,3}

\addinstitution{
	Institute of Imaging \& Computer Vision\\
	RWTH Aachen University\\
	Aachen, Germany
}
\addinstitution{
	Faculty of Informatics and Data Science, University of Regensburg, Germany
}
\addinstitution{
	Fraunhofer Institute for Digital Medicine MEVIS, Bremen, Germany
}

\runninghead{Long Chen et al.}{Instance Segmention via Layering}

\begin{document}
	
	\maketitle
	
	\begin{abstract}
		Instance segmentation aims to delineate each individual object of interest in an image. 
		State-of-the-art approaches achieve this goal by either partitioning semantic segmentations or refining coarse representations of detected objects. 
		In this work, we propose a novel approach to solve the problem via object layering, i.e. by distributing crowded, even overlapping objects into different layers. By grouping spatially separated objects in the same layer, instances can be effortlessly isolated by extracting connected components in each layer. 
		In comparison to previous methods, our approach is not affected by complex object shapes or object overlaps.  
		With minimal post-processing, our method yields very competitive results on a diverse line of datasets: \textit{C. elegans} (BBBC), Overlapping Cervical Cells (OCC) and cultured neuroblastoma cells (CCDB). The source code is publicly available~\footnote{https://github.com/looooongChen/instSeg/}.
	\end{abstract}
	
	\section{Introduction}
	\label{sec:intro}
	Different from semantic segmentation, which pays no attention to the individual objects, instance segmentation aims not only to associate every pixel of an image with a class label but also to delineate objects of the same class as individuals. This task becomes more challenging, when objects are densely located or even overlapping with each other.
	
	A prevalent top-down class of instance segmentation methods is detection-based, which firstly localize an instance with a coarse shape representation and then refine the shape in an additional step. As a paradigm, Mask-RCNN~\cite{mrcnn} refines bounding boxes obtained from Region-based Convolutional Neural Networks (R-CNN)~\cite{rcnn,frcnn}. Relying on non-maximum suppression (NMS) to remove duplicate predictions, detection-based methods become less competent in the cases of dense clusters and overlapping objects. A finer polygon representation approach was proposed by~\cite{StarDist} to reduce false suppressions. However, NMS has a methodological flaw for objects that inherently overlap. Moreover, many objects, especially in the biomedical domain, cannot be well-approximated by tractable shape representations, such as bounding boxes and star-convex polygons.   
	
	Free from false suppression and coarse shape approximation, alternative bottom-up approaches obtain instances by grouping pixels~\cite{dcan,ssap,deepWatershed}. DCAN~\cite{dcan} predicts the object boundary explicitly and groups connected pixels that are separated by these boundaries as instances. However, this approach is sensitive to broken boundaries. A few misclassified pixels can lead to erroneous merging of adjacent objects. Graph partition based on pixel-pair affinity~\cite{ssap} and the watershed transform~\cite{deepWatershed} are more robust in terms of grouping, but the pipeline as a whole relies heavily on post-processing, with the learning model only optimized with intermediate results. In addition, grouping-based approaches are inherently incapable of handling overlapping objects, where one pixel may belong to more than one object.
	
	Recent research~\cite{disEmb,cosEmb,rnnEmb,impemb} introduces pixel embedding for grouping. In these approaches, a deep neural network is trained to map pixels into an embedding space, in which pixel embeddings from the same object are close, while those from different objects, especially adjacent objects, are apart. Then, pixels are grouped in the embedding space with low-level clustering algorithms, such as Mean Shift~\cite{meanShift} and DBSCAN~\cite{DBSCAN}. Although pixel embedding based grouping has proven to be more robust, it still can not avoid the shortcomings of grouping-based methods: the reliance on post-processing, the optimization of intermediate results, and the incapability to segment overlapping objects. 
	
\begin{figure}[h]
		\centering
		\includegraphics[width=\textwidth]{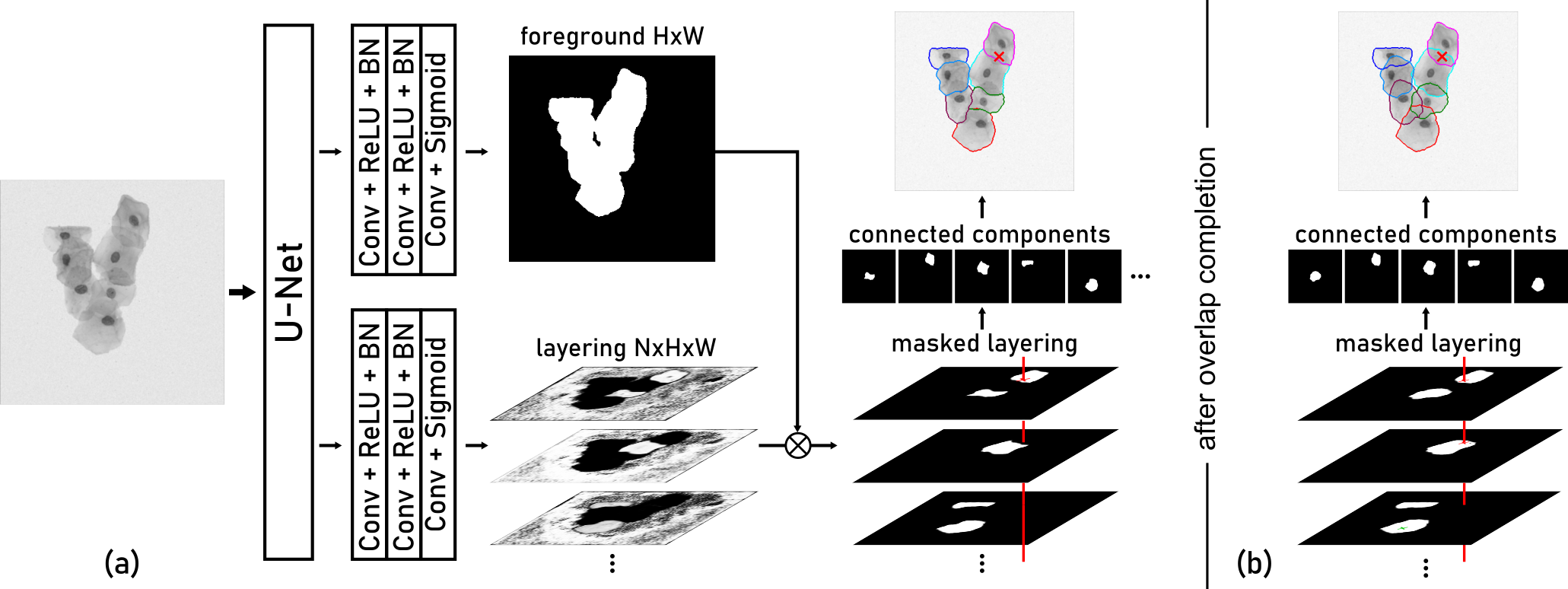}
		
		\caption{Our approach eliminates the object crowding and overlapping by distributing adjacent objects into different layers. (a) To achieve this, each foreground pixel is assigned by drawing pixels of the same object into the same layer and pushing pixels of adjacent objects into different layers (Section~\ref{sec:layering}). Areas where overlap occurs are ignored in this training phase. (b) After object layering converges, overlapping areas are trained to complete the object (Section~\ref{sec：completion}). At locations where overlap occurs (the red cross and line), more than one layer will give foreground prediction.}
		\label{fig:overview}
	\end{figure}

	Our work is partially inspired by the pixel embedding approach. To alleviate the aforementioned limitations, we propose to train a more structured embedding space. Specifically, we restrict one object to "live" in one dimension. Correspondingly, embedding vectors will be one-hot in overlap-free areas and have more than one active digits at locations where an overlap of objects occurs (Figure~\ref{fig:overview}). 
	We figuratively call this process of objects being distributed to different dimensions "layering", and use the terms "dimension" and "layer" interchangeably in the following context.
	Since our training loss penalizes adjacent objects having the same embedding vector, only spatially separated objects can be assigned to the same layer, eliminate crowding and overlapping of objects. Therefore, instances can be effortlessly obtained by finding connected components in each layer.
	
	The main contributions of our work are as follows: We propose (1) a novel approach to eliminate crowdedness and overlap of instances by layering objects into different output layers; (2) an approach of \emph{spontaneous} object layering through deep model learning; (3) a concise and effective framework for segmentation of densely distributed objects without data-specific post-processing efforts. 
	
	To our best knowledge, our work is the first pixel embedding based approach that does not require explicit pixel clustering, and is capable of handling object overlap. Competing with several state-of-the-art approaches, our method yields comparable or better results on a diverse line of datasets: \textit{C. elegans}~\cite{bbbc} (BBBC), overlapping cervical cells~\cite{occ14-B,occ14-A} (OCC) and cultured neuroblastoma cells~\cite{ccdb} (CCDB).
	
	\section{Proposed Method}
	\subsection{Overview}
	\label{sec:overview}
	Our model consists of two output branches: the foreground branch and the layering branch (Figure~\ref{fig:overview}). The foreground segmentation, as an auxiliary, excludes background pixels from further processing, while the layering branch is devoted to separating foreground instances.
	
	The foreground branch uses a 1-channel convolutional layer with \emph{Sigmoid} activation for output. The output layer can also be set to multi-channel and \emph{Softmax} activated in the case of more than one semantic category. The layering branch has $N$ output channels with the \emph{Sigmoid} activation, each of which is trained to only contain spatially separated objects. Since objects in the same layer exhibit no contact or overlap, they can be effortlessly isolated by extracting connected components. 
	
	We train the model in two phases: layering training and overlap completion. In the first phase, under the supervision of our proposed layering loss (Section~\ref{sec:layering}), the model learns to assign foreground pixels to one of the layers, maintaining the restriction that neighboring objects should be located in different layers. We only consider image areas where no objects overlap in this phase, since the layering loss exclusively chooses one layer for one pixel. After the layering training converges, we generate $N$ binary masks by ordering each object mask into one of the $N$ layers, based on the layering results of the object's overlap-free part. It is worth mentioning that more than one layer will be positive at locations where an overlap of objects occurs. A \emph{Dice}-like loss (Section~\ref{sec：completion}) is then computed with the generated masks and further included in the second training phase. The overlapping area is explicitly trained in this phase, to complete the intact object.
	
	In summary, the model is trained in two phases with the following loss:
	\begin{equation}
		L = [L_{foreground}]_S^{1,2} + [L_{layering}]_{S_{fn}}^{1,2} + [L_{overlap}]_{S_f}^2,   
	\end{equation}
	where the subscript of $[\cdot]$ denotes on which area a loss term is computed: $S$, $S_f$ and $S_{fn}$ represent three progressively smaller areas, namely, the whole image, the foreground and the foreground without object overlap. The superscript indicates in which training phase the loss term is included. The standard \emph{Crossentropy} is used as foreground training loss $L_{foreground}$. Details of $L_{layering}$ and $L_{overlap}$ are depicted in the following two sections.

	\subsection{Layering Loss}
	\label{sec:layering}
	The object layering is achieved by drawing pixels of the same object together into the same dimension with an attracting loss term $L_{attr}$ and pushing away neighboring objects into different dimensions with a repelling loss term $L_{rep}$. Our loss is constructed aroud the cosine similarity
	$D(\mathbf{e_i},\mathbf{e_j}) = \frac{\mathbf{e_i}^\top \mathbf{e_j}}{\|\mathbf{e_i}\|_2\|\mathbf{e_j}\|_2}$~\cite{cosEmb,impemb}, which is zero when two vectors are located in two othogonal spaces. The operation  $\|\cdot\|_2$ computes the \(\mathcal{L}^2\) norm.
	
	We push adjacent objects into the orthogonal space of each other, while non-adjacent objects can stay in the same cluster. Assuming that there are $C$ objects ($\{O_i | i = 1,2,...,C\}$) in an image, we represent the overlapping and overlap-free part of an object with $O_i^o$ and $O_i^n$, respectively. The attracting and repelling terms can thus be formulated as:
	
	\begin{align}
		L_{attr} = & 1 - \frac{1}{\sum_{i=1}^{C}|O_i^n|} \sum_{i=1}^{C} \sum_{p \in O_i^n} D(\mathbf{e_p}, \mathbf{u_i}) ^ 2\;, \\
		L_{rep} = &  \frac{1}{C} \sum_{i=1}^{C} \frac{1}{|Adj(O_i)|}\sum_{j\in Adj(O_i)} D(\mathbf{u_i}, \mathbf{u_j}) ^2 \;,
	\end{align}
	
	\noindent where $e_p$ indicates the embedding vector of pixel $p$ and $\mathbf{u_i} = \frac{1}{|O_i^n|} \sum_{p \in O_i^n}\mathbf{e_p} $ is the mean embedding of overlap-free part of the $i$-th object. $Adj(O_i)$ represents the set of adjacent objects to object $O_i$, whose shortest distances to $O_i$ are less than a threshold $t$ ($t=15$ pixels in this work). The operator $|\cdot|$ returns the element number of a set. For example, $|O_i^n|$ is the pixel number of area $O_i^n$ and $|Adj(O_i)|$ is the number of adjacencies of object $O_i$.
	
\begin{figure}[h]
		\centering
		\includegraphics[width=\textwidth]{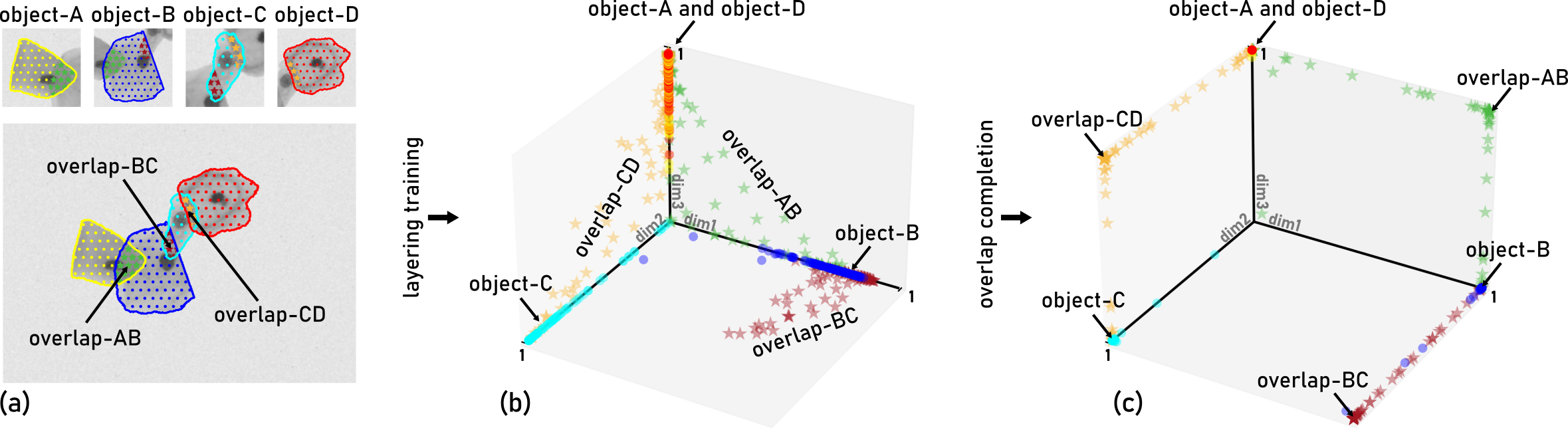}
		
		\caption{Visualization of trained embedding space. (a) Four objects with three overlapping areas are present in this example. (b) After layering training, the object parts that do not overlap are ideally distributed in different dimensions (layers). Spatially separated objects, such as A and D, can exist in the same dimension. Pixels of overlapping areas, untrained in this phase, live randomly in the plane composed of axes to which the overlapping objects belong. (c) After overlap completion, pixels of the overlapping area will congregate to the point $(1,1)$, indicating a foreground area in both layers. And, pixels of the non-overlap part are also more compactly distributed and closer to 1.}
		\label{fig:embedding}
	\end{figure}

	Training with the attracting term $L_{attr}$ and the repelling term $L_{rep}$ specifies embedding vectors to locate in mutually orthogonal spaces, but it does not guarantee that the vectors only "live" in one layer (dimension), i.e. they are distributed on the standard axes (see Figure~\ref{fig:embedding}). Therefore, we introduce the sparse term $L_{sparse}$, which imposes preferences for the vector whose digits, except for a single one, are suppressed to 0:
	\begin{align}
		L_{sparse} =& 1 - \frac{1}{\sum_{i=1}^{C}|O_i^n|} \sum_{i=1}^{C} \sum_{p \in O_i^n} \max (\frac{\mathbf{e_p}}{||\mathbf{e_p}||_2})\;,
	\end{align}
	where the operator $\max(\cdot)$ takes the value of the maximal digit. 
	
	The layering loss consists of the three terms above:
	\begin{align}
		L_{layering} = L_{attr} + L_{rep} + \lambda L_{sparse}\;,
	\end{align}
	where $\lambda$ is a weighting constant, and we use $\lambda=0.1$ in this work. It is worth mentioning that only the overlap-free foreground part is involved in the calculation of $L_{layering}$.

	\subsection{Overlap Completion}
	\label{sec：completion}
	After layering training, the overlap-free parts of objects are assigned to one of the layers, while the overlapping parts remain untrained. Generally, the untrained pixels from overlapping areas will randomly belong to one dimension to which one of the objects that overlap belongs (Figure~\ref{fig:embedding}b). Accordingly, a random boundary between objects will be formed, as illustrated in Figure~\ref{fig:overview}a.
	
	To train the region with overlap, we firstly generate $N$ binary masks, denoted by $S \in \{0,1\}^{H \times W \times N}$, by placing the silhouette of each object into one of the $N$ masks based on the layering results of the object's overlap-free part. Indexing the stack of N layers at different pixel positions $p$ with $S_p\in \{0,1\}^N$, we define the following \emph{Dice}-like~\cite{dice} loss for overlap completion training: 
	
	\begin{align}
		L_{overlap} =& 1 - 2\frac{\sum_{i=1}^{C} \sum_{p \in O_i} \mathbf{e_p}^\top \mathbf{S_p}}{\sum_{i=1}^{C} \sum_{p \in O_i} (\mathbf{1}^\top \mathbf{e_p} + \mathbf{1}^\top \mathbf{S_p}) } \;.
	\end{align}
	
	The dot product between binary vector $\mathbf{e_p}$ and $\mathbf{S_p}$ on the numerator of $L_{overlap}$ is analogous to the intersection in \emph{Dice} loss, whose value is maximized when $\mathbf{e_p}$ and $\mathbf{S_p}$ are exactly the same. The denominator sums up all digits of vector $\mathbf{e_p}$ and $\mathbf{S_p}$, which is equivalent to the sum in the \emph{Dice} loss. The overlap completion loss $L_{overlap}$ is applied to the foreground, including the overlapping and non-overlapping areas. The vector $\mathbf{S_p}$ is one-hot in areas without overlap, while it has more than one non-zero digits at locations where overlap occurs.

	\subsection{Post-Processing}
	\label{sec:post-processing}
	Since each of the $N$ layers only contains spatially separated objects, high-quality instance segmentations can be obtained with minimal post-processing effort, only involving simple and computationally efficient operations. Our post-processing steps are also general for all instance segmentation tasks, without utilizing any prior knowledge specific to a certain dataset. Detailed steps are listed in Algorithm~\ref{alg:postprocessing}. It worth mentioning that line 7-8 can be ignored if there is no object overlap for certain tasks.
	
	The post-processing requires two trivial parameters: a threshold $\tau$ and a minimal object size $S_{min}$. The value $\tau$ is used for foreground thresholding (lines 1 and 7 in Algorithm~\ref{alg:postprocessing}). The minimal object size $S_{min}$ is responsible for eliminating small noisy objects (lines 2 and 11 in Algorithm~\ref{alg:postprocessing}). We use $\tau=0.5$ and $S_{min} = 250$ for all experiments in this work.  
	
	\begin{algorithm} \caption{Post-processing to obtain object segmentations}
		\label{alg:postprocessing}
		\begin{algorithmic}[1]
			\Require (raw prediction) foreground $F_{raw}\in (0,1)^{H \times W}$, layering $L_{raw}\in (0,1)^{H \times W \times N}$
			\Require (parameters) threshold $\tau$, minimal object size $S_{min}$
			\State threshold $F_{raw}$ with value $\tau$ to get foreground $F$
			\State remove connected components in $F$ that are smaller than $S_{min}$
			\State initial $N$ empty binary layering masks $L\in \{0,1\}^{H \times W \times N}$
			\For{each pixel location $ i \in \{1,2,...,H\},j \in \{1,2,...W\}$ and layer $ k \in \{1,2,...,N\}$}
			\If {$L_{raw}(i,j,k)$ is the largest in $L_{raw}(i,j,:)$}
			\State set $L(i,j,k)$ to 1
			\ElsIf  {$L_{raw}(i,j,k)$ is larger than $\tau$} \Comment{omit if no overlap exists}
			\State set $L(i,j,k)$ to 1 \Comment{omit if no overlap exists}
			\EndIf
			\EndFor
			\State take all connected components larger than $S_{min}$ in each layer as objects
		\end{algorithmic}
	\end{algorithm}

	\section{Experiments and Results}
	
	\subsection{Implementation}
	\label{sec:implementation}
	
	While our approach is not tied to a particular network architecture, we perform experiments with UNet~\cite{unet}. In our implementation, we apply batch normalization after each convolutional layer. In addition, we experiment with a UNet-S and a UNet-L, which have, respectively, one less and one more convolutional block on the contracting and expansive path than the original UNet (denoted as UNet-M), to investigate the effect of receptive field size.
	
	The foreground branch and the layering branch share the same feature map from the UNet backbone. Two 3x3 covolutional layers with 64 features are added before the output layer to avoid "feature conflict". We use an $N$-channel convolutional layer ($N=8$ in this work) and a $1$-channel convolutional layer with the \emph{sigmoid} activation for the final output. 
	
	The model was trained with \emph{RMSprop} optimizer~\cite{rmsprop} with a learning rate of $1e^{-4}$ exponentially decayed to 0.9 every 10000 steps. 
	The training dataset was randomly split into 90$\%$ and 10$\%$ for training and validation.
	The layering training phase lasted 1000 epoches. Then, the overlap completion training continued from the best model of layering training for another 500 epoches. The best model of overlap completion training was used as the final model for evaluation. The "best" models are chosen based on validation results.
	
	\subsection{Datasets and Evaluation Metrics}
	\label{sec:data}
	For the evaluation, we use three biomedical image datasets containing a population of objects with different shapes, different degrees of density and overlap. All datasets were augmented using random horizontal and vertical flips, random rotation, random gamma $\gamma \in (0.5, 2)$ correction transform and elastic deformation.
	
	\noindent\textbf{BBBC010} contains 100 bright-field microscopic images of live/dead \textit{C. elegans} ~\cite{bbbc}. The \textit{C. elegans} are slender, bilaterally symmetrical objects in curved or ring-shaped poses, which may cross over others in this data set. Subset D was left for evaluation and the rest for training. We crop the image and a 448x448 pixel area containing objects remains.

	\noindent\textbf{OCC2014} is an EDF (extended depth of field) image collection of overlapping cervical cells from Pap smears~\cite{occ14-B,occ14-A}, consisting of 16 real and 945 synthesized images. 
	The cells are roundish and densely clustered, and the overlap can be larger than 50$\%$ of the object size. 
	In our experiments, the original training set and half of the test set served for training. Evaluation was conducted on the other half.
	The images are resized to 320x320 pixels.
	
	\noindent\textbf{CCDB6843} contains 100 wide field fluorescent images of cultured neuroblastoma cells collected by~\cite{ccdb}. The cells are densely located and irregularly shaped. We randomly chose 24 images as the test set. All images are resized to 448x448 pixels.
	
	In our evaluation, a predicted object $I_{pred}$ is considered to be a successful match (true positive $TP_t$) if its intersection over union $IoU=\frac{I_{pred} \cap I_{gt}}{I_{pred} \cup I_{gt}}$ with a ground truth object is larger than a given threshold $t$, while unmatched predictions and ground truth objects are counted as false positive ($FP_t$) and false negative ($FN_t$). Using these values, a measure of detection accuracy can be calculated: $AP_t=\frac{TP_t}{TP_t+FP_t+FN_t}$. By passing from loose to strict thresholds, the segmentation accuracy is also reflected. To better quantify the segmentation performance, we also calculate the Aggregated Jaccard Index ($AJI$)~\cite{aji}.
	
	\begin{figure}[]
		\centering
		
		\begin{minipage}[b]{0.119\textwidth}
			\centering
			Layer-1 
			\includegraphics[width=\linewidth]{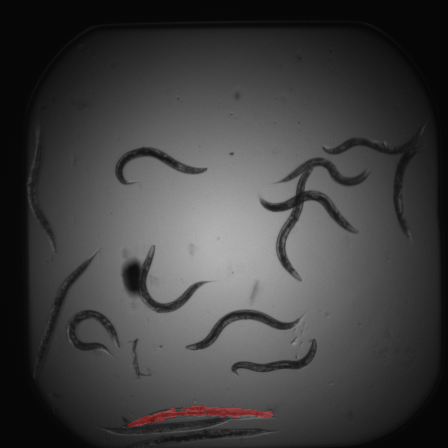} 
		\end{minipage}
		\begin{minipage}[b]{0.119\textwidth}
			\centering
			Layer-2
			\includegraphics[width=\linewidth]{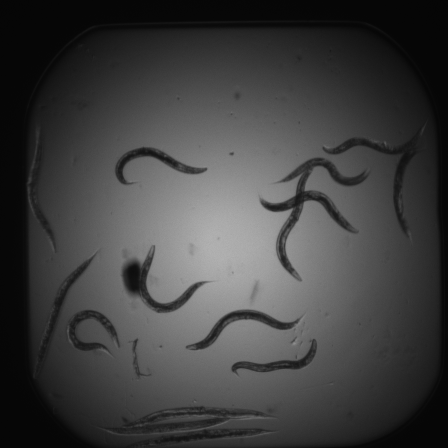} 
		\end{minipage}
		\begin{minipage}[b]{0.119\textwidth}
			\centering
			Layer-3
			\includegraphics[width=\linewidth]{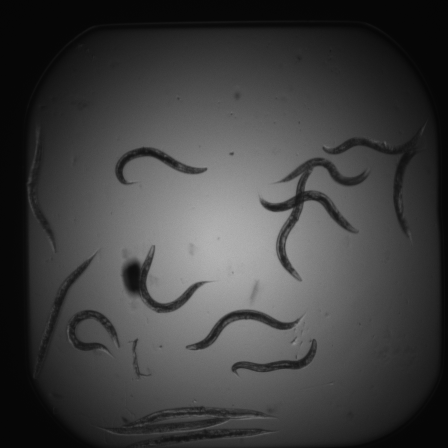} 
		\end{minipage} 
		\begin{minipage}[b]{0.119\textwidth}
			\centering
			Layer-4
			\includegraphics[width=\linewidth]{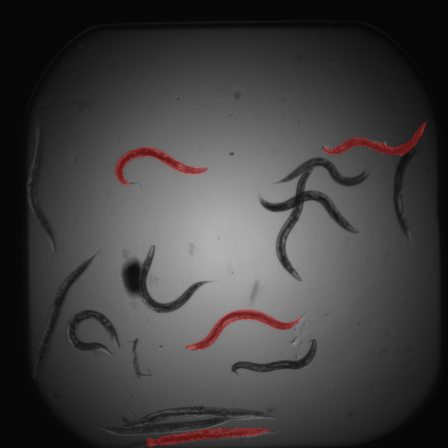} 
		\end{minipage} 
		\begin{minipage}[b]{0.119\textwidth}
			\centering
			Layer-5
			\includegraphics[width=\linewidth]{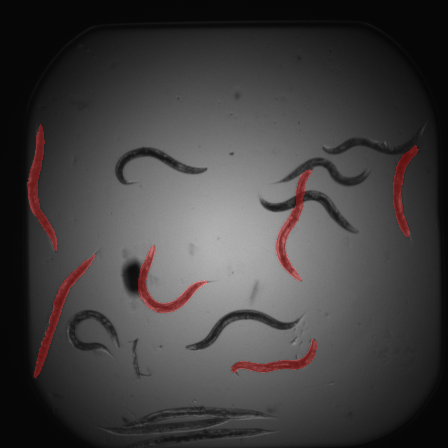} 
		\end{minipage} 
		\begin{minipage}[b]{0.119\textwidth}
			\centering
			Layer-6
			\includegraphics[width=\linewidth]{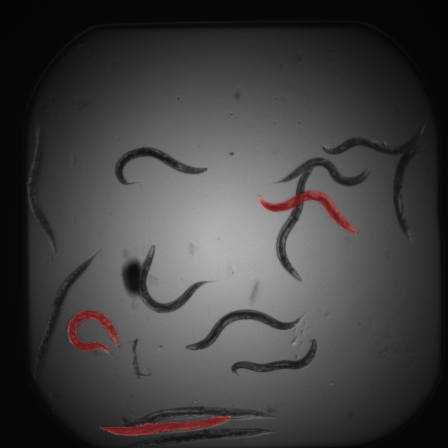} 
		\end{minipage}  
		\begin{minipage}[b]{0.119\textwidth}
			\centering
			Layer-7
			\includegraphics[width=\linewidth]{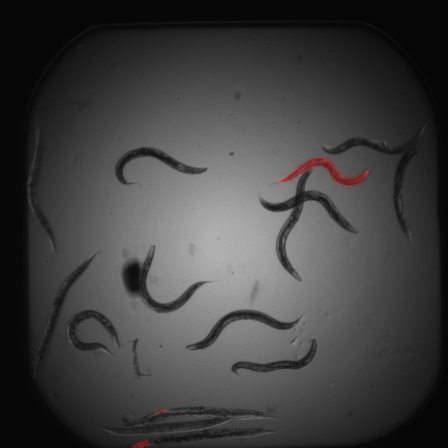} 
		\end{minipage}
		\begin{minipage}[b]{0.119\textwidth}
			\centering
			Layer-8
			\includegraphics[width=\linewidth]{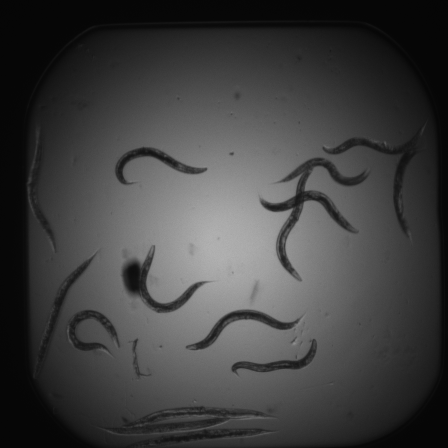} 
		\end{minipage}    
		
		\includegraphics[width=.119\textwidth]{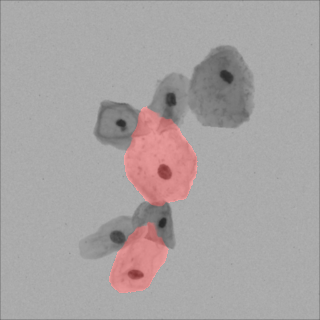}
		\includegraphics[width=.119\textwidth]{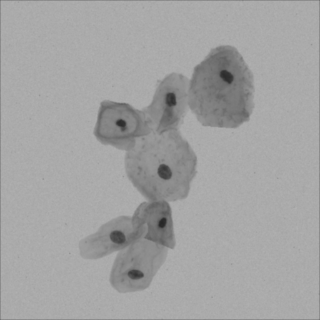}
		\includegraphics[width=.119\textwidth]{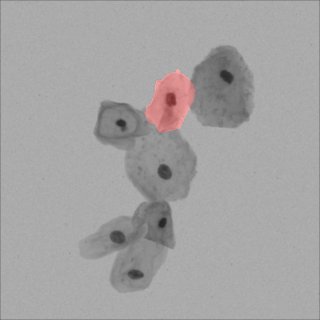}
		\includegraphics[width=.119\textwidth]{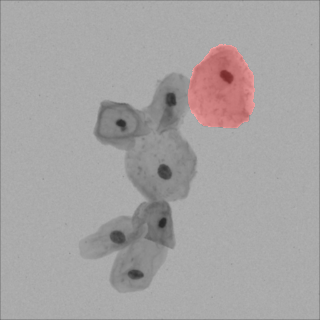}
		\includegraphics[width=.119\textwidth]{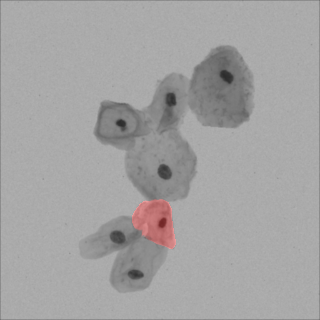}
		\includegraphics[width=.119\textwidth]{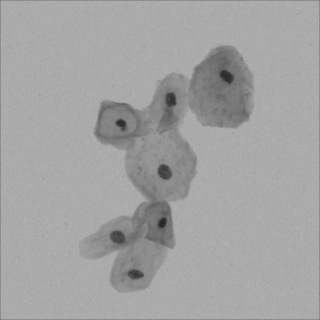}
		\includegraphics[width=.119\textwidth]{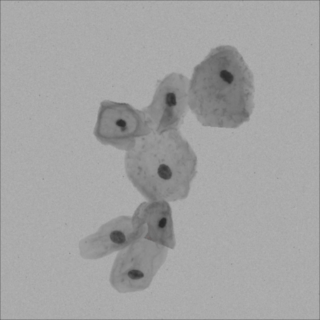}
		\includegraphics[width=.119\textwidth]{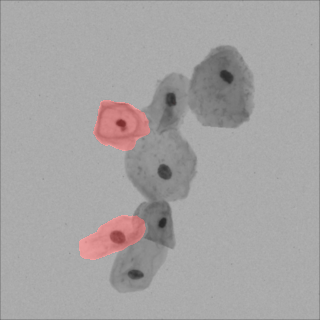}
		
		\includegraphics[width=.119\textwidth]{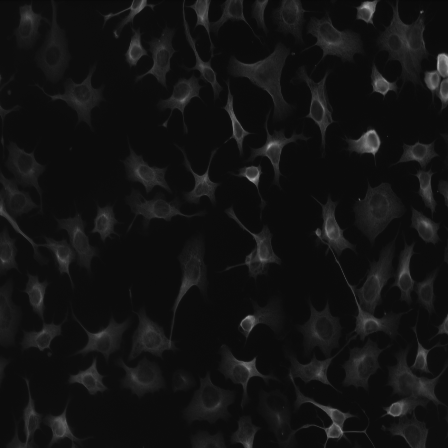}
		\includegraphics[width=.119\textwidth]{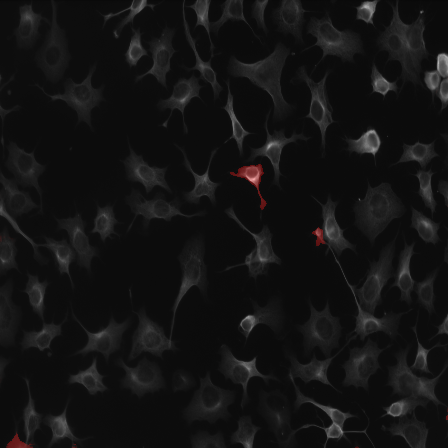}
		\includegraphics[width=.119\textwidth]{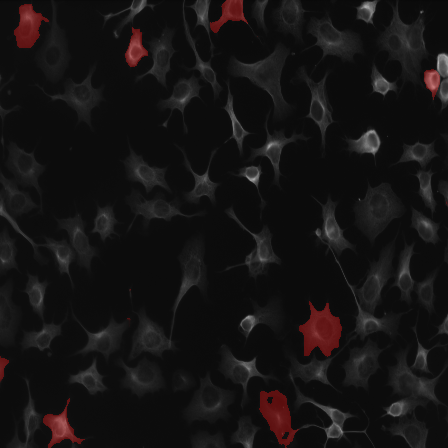}
		\includegraphics[width=.119\textwidth]{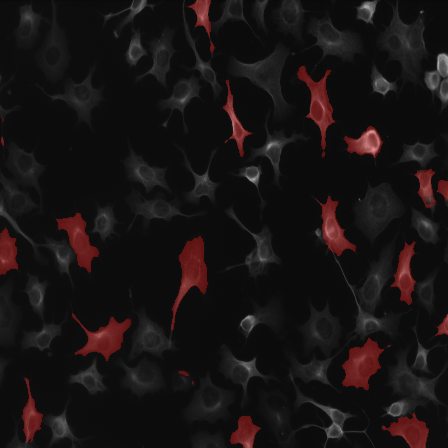}
		\includegraphics[width=.119\textwidth]{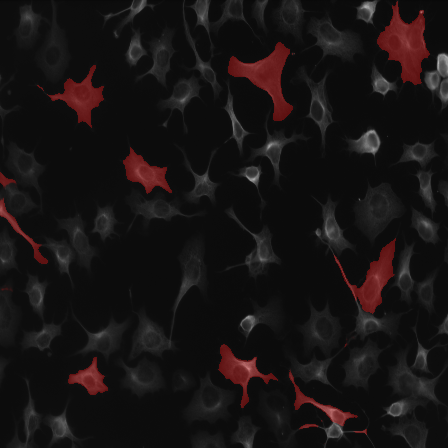}
		\includegraphics[width=.119\textwidth]{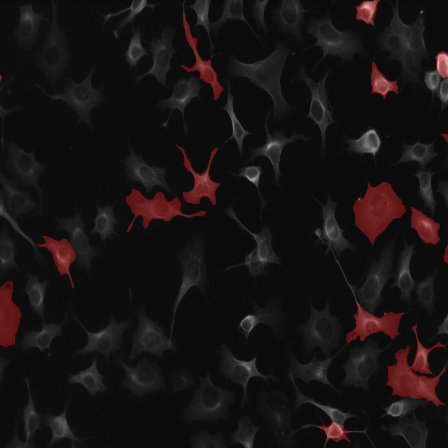}
		\includegraphics[width=.119\textwidth]{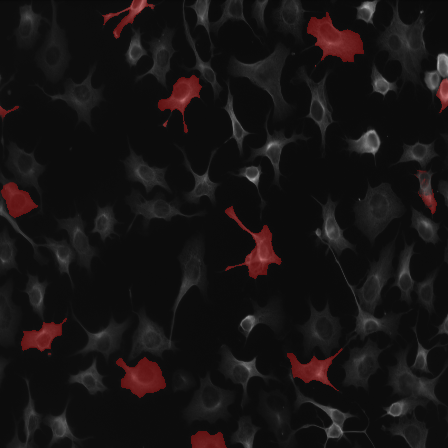}
		\includegraphics[width=.119\textwidth]{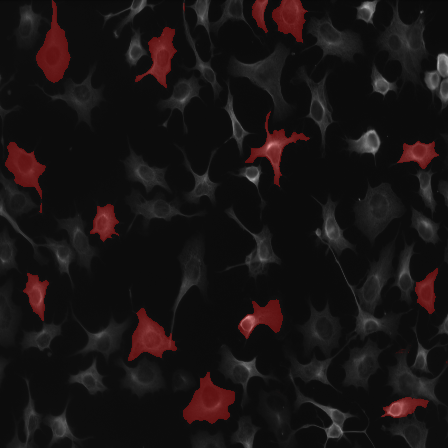}	
		
		\caption{Activation map of each layer: results on the datasets BBBC, OCC and CCDB from top to bottom. Spatially adjacent objects are distributed to different layers, eliminating crowding and overlapping of objects. Meanwhile, not all layers contain objects, which suggests that, by keeping only adjacent objects separate, layers are fully utilized.}
		\label{fig:layering}
	\end{figure}

	\begin{figure}[]
		\centering
		\begin{minipage}[b]{0.135\textwidth}
			\centering
			Image
			\includegraphics[width=\linewidth]{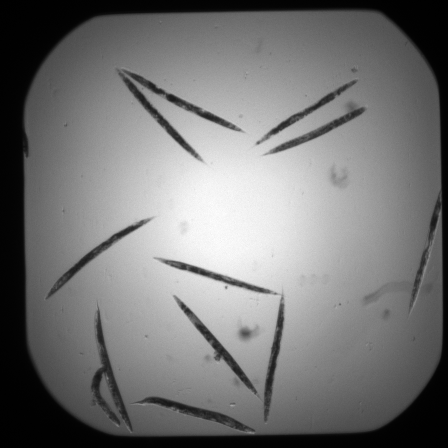} 
		\end{minipage}
		\begin{minipage}[b]{0.135\textwidth}
			\centering
			UNet-2
			\includegraphics[width=\linewidth]{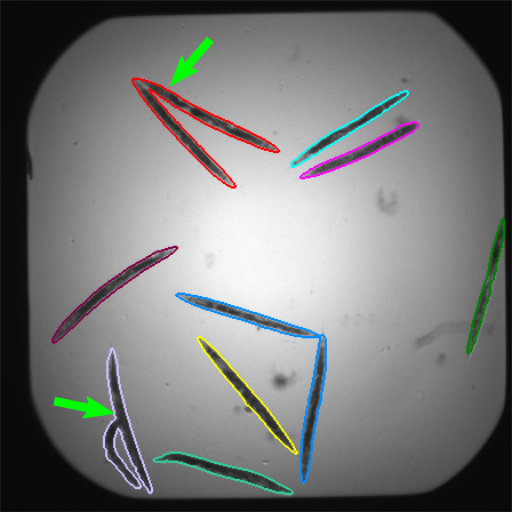} 
		\end{minipage} 
		\begin{minipage}[b]{0.135\textwidth}
			\centering
			UNet-3
			\includegraphics[width=\linewidth]{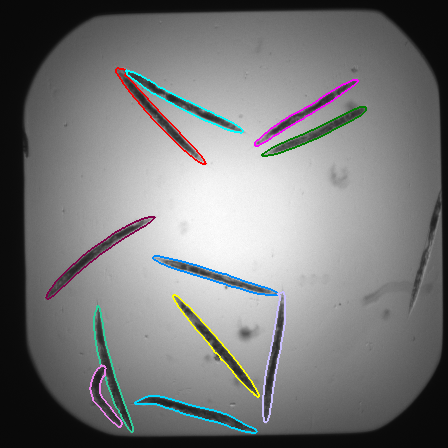} 
		\end{minipage} 
		\begin{minipage}[b]{0.135\textwidth}
			\centering
			M-RCNN
			\includegraphics[width=\linewidth]{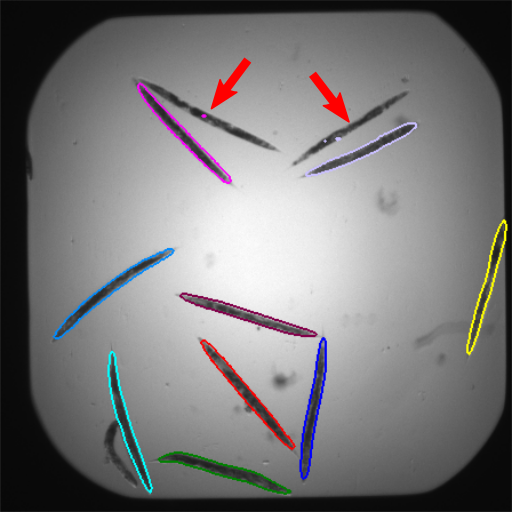} 
		\end{minipage} 
		\begin{minipage}[b]{0.135\textwidth}
			\centering
			StarDist
			\includegraphics[width=\linewidth]{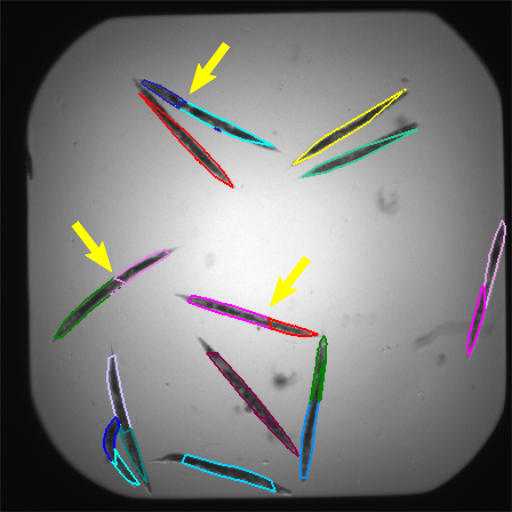} 
		\end{minipage}  
		\begin{minipage}[b]{0.135\textwidth}
			\centering
			Ours
			\includegraphics[width=\linewidth]{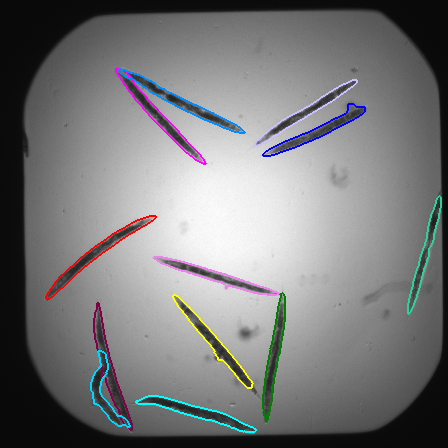} 
		\end{minipage}  
		\begin{minipage}[b]{0.135\textwidth}
			\centering
			GT
			\includegraphics[width=\linewidth]{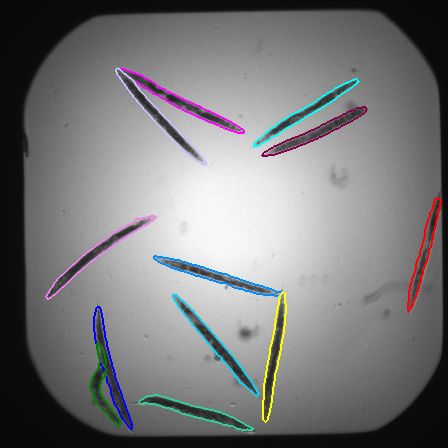} 
		\end{minipage}

		\includegraphics[width=.135\textwidth]{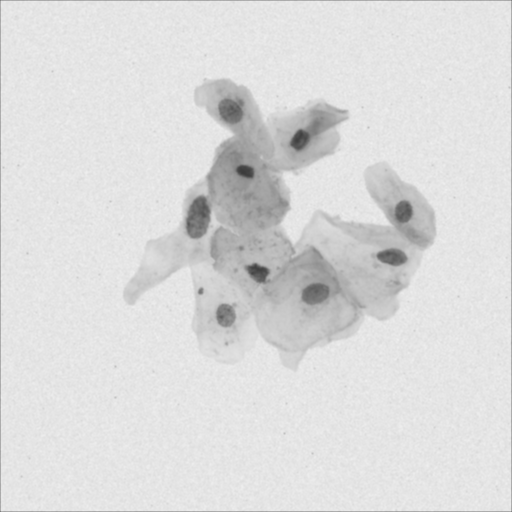}
		\includegraphics[width=.135\textwidth]{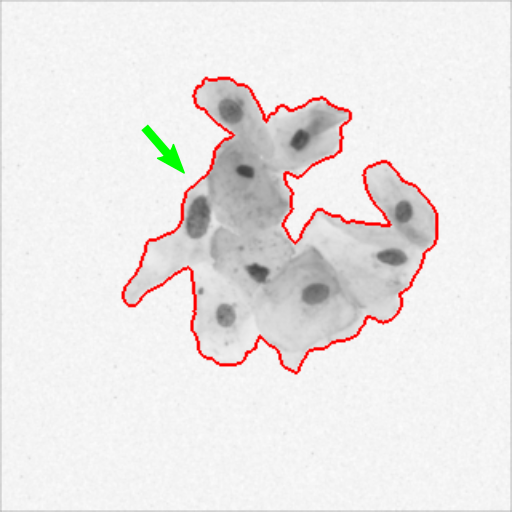}
		\includegraphics[width=.135\textwidth]{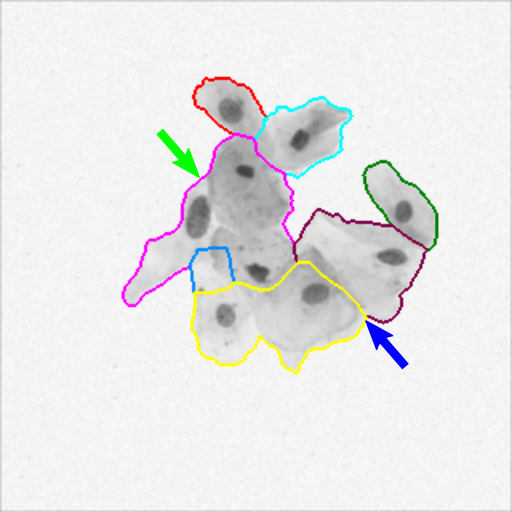}
		\includegraphics[width=.135\textwidth]{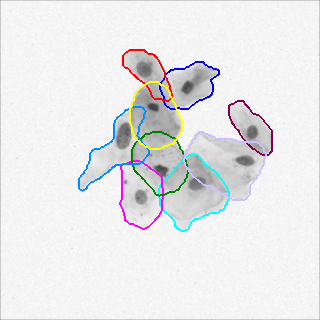}
		\includegraphics[width=.135\textwidth]{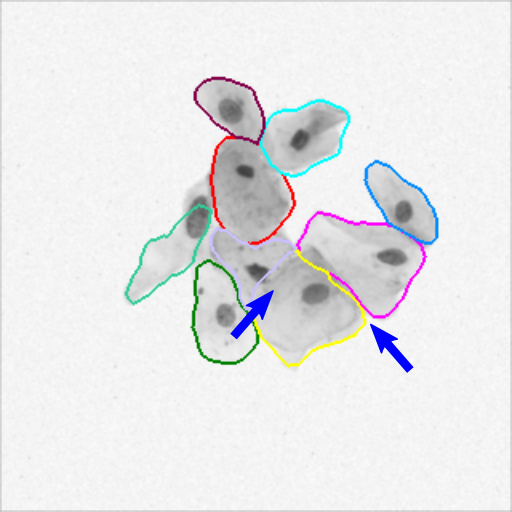}
		\includegraphics[width=.135\textwidth]{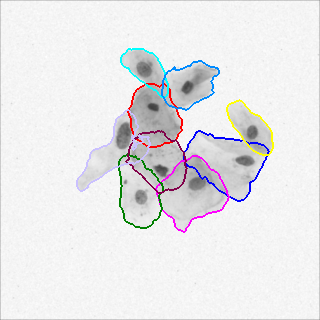}
		\includegraphics[width=.135\textwidth]{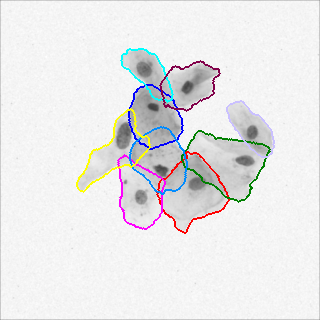}
		
		\includegraphics[width=.135\textwidth]{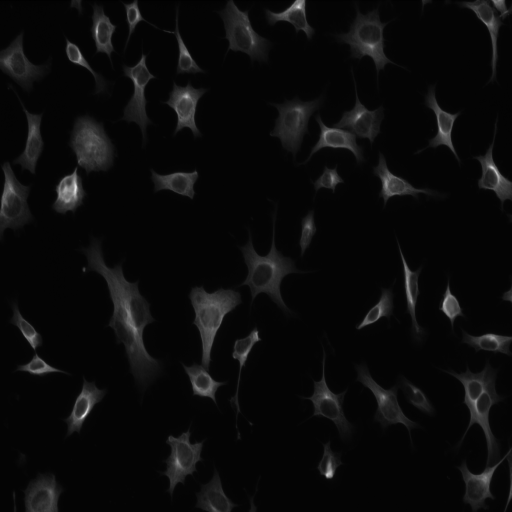}
		\includegraphics[width=.135\textwidth]{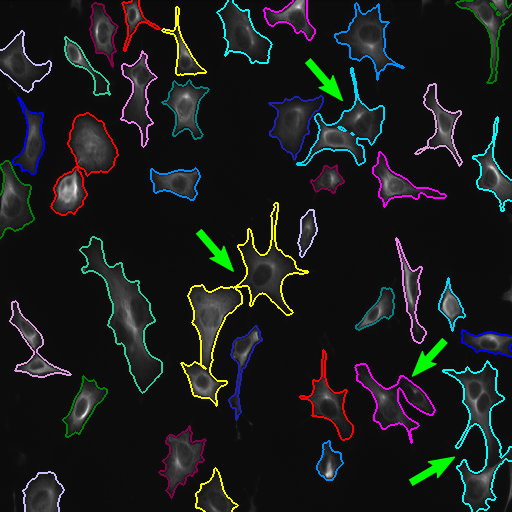}
		\includegraphics[width=.135\textwidth]{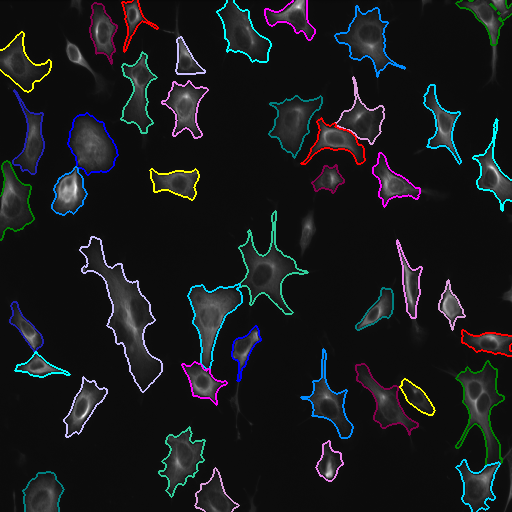}
		\includegraphics[width=.135\textwidth]{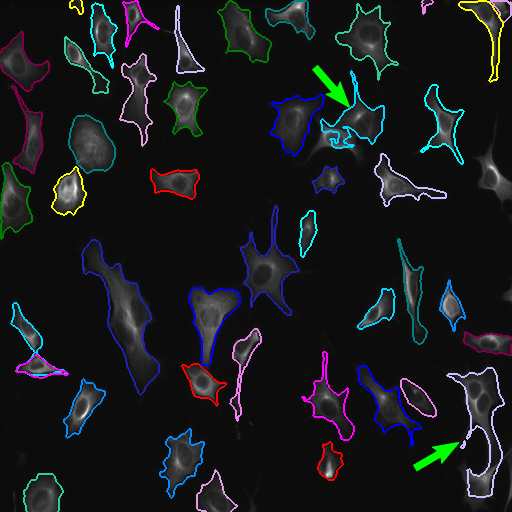}
		\includegraphics[width=.135\textwidth]{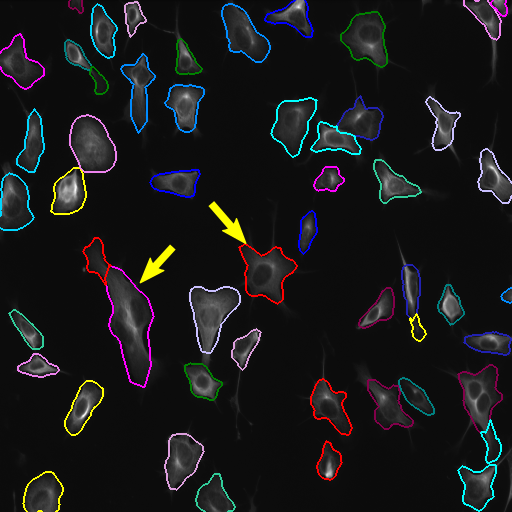}
		\includegraphics[width=.135\textwidth]{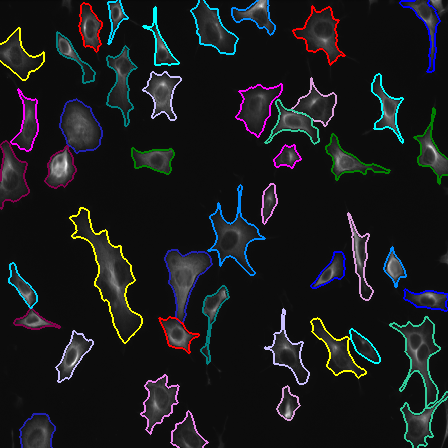}
		\includegraphics[width=.135\textwidth]{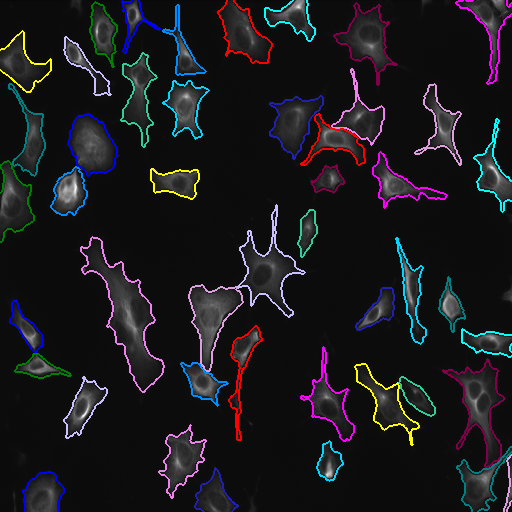}
		
		\caption{Qualitative segmentation results on the datasets BBBC, OCC and CCDB from top to bottom. A few typical errors are marked with arrows: false suppression (red), merged objects (green), inaccurate shape (yellow) and incapability to handle overlap (blue).}
		\label{fig:seg}
	\end{figure}
	
	\subsection{Competing Methods}
	\label{sec:baselines}
	
	\noindent\textbf{UNet-2/3:} We train UNet~\cite{unet} models as semantic segmentation tasks with a two labels (object and background) and a three labels (object, boundary and background) setup. In terms of network structure, UNet is the least different from our model: UNet output all objects with a single image plane, whereas our model has several containing layered objects. 
	
	\noindent\textbf{Mask-RCNN:} Mask-RCNN~\cite{mrcnn} localizes objects by proposal classification and non-max suppression (NMS). Then, segmentation is performed on each object bounding box. The NMS threshold was set to 0.7 for all experiments.
	
	\noindent\textbf{StarDist:} Without an explicit segmentation step, StarDist~\cite{StarDist} obtains object masks by combining distances from the center to the boundary along different directions. We used 32 radial directions in our experiments. 
	
	All methods except Mask-RCNN were trained from scratch, while the Mask-RCNN model was fine-tuned on a COCO pretrained model~\cite{coco} with the ResNet-101 backbone~\cite{resnet}. For the 3-label UNet and StarDist, we used a pseudo-boundary as an approximated separation of overlapping regions, obtained by skeletonizing~\cite{skeleton} the binary mask of overlapping regions together with boundaries.

	\subsection{Results and Discussion}
	\label{sec:results}

	
	For a careful interpretation, we discuss the methods from the following aspects: the handling of object shape, touching objects, object overlap and the effect of high object density. 
	
	\noindent\textbf{Shape:} Using radial directions, StarDist has difficulty in reconstructing slender shapes and boundaries with fine structure. This is reflected by the broken body of \textit{C. elegans} and the roundish approximation of the neuroblastoma cells (Figure~\ref{fig:seg}). Other methods, which perform pixel-wise segmentation, are less affected. 
	
	\noindent\textbf{Touching objects:} Since it does not take clustered objects explicitly into consideration, the 2-label UNet performs suboptimally on all datasets due to false fusion,
	while others are object-aware to different degrees under different conditions.
	
	\noindent\textbf{Object overlap:} Since UNet and StarDist naturally assign one pixel to one object, an overlapping area can only be handled with approximated boundaries. In case of less severe overlap, such as on the BBBC and CCDB datasets, UNet (3-label) still achieves very good results, although its performances drops significantly when the overlap ratio increases on the dataset OCC. 
	
	\noindent\textbf{Object density:} Mask-RCNN suffers from the false suppression of NMS, when the bounding boxes of two objects overlap with a large ratio, as, for example, in the case of the two parallel and close \textit{C. elegans} in Figure~\ref{fig:seg}. We also find that Mask-RCNN has difficulty in distinguishing two closely located irregular-shaped objects, such as the neuroblastoma cells in Figure~\ref{fig:seg}.
	
	As described above, segmentation errors, including false suppression, merged objects, inaccurate shape and the incapability to handle overlap (Figure~\ref{fig:seg}), occur on different approaches, depending on the data characteristics. By contrast, our method is more robust in all of these aspects, and, therefore, achieves the best and, evidently, better results on the dataset BBBC and CCDB (Table~\ref{tab:resulsts}). On the dataset OCC, the performance of our approach is only marginally worse than Mask-RCNN in terms of the Aggregated Jaccard Index ($AJI$).
	
	\begin{table}
		\centering
		\caption{Quantitative evaluation. Average precisions ($AP_t$) under different $IoU$ thresholds, mean values of average precision $AP$ over these thresholds, and the Aggregated Jaccard Index ($AJI$) are reported. The best two results are shown in bold, and the best is underlined.}
		\label{tab:resulsts}
		\begin{tabular}{c|c|ccccccc}
			\hline
			\multicolumn{2}{c|}{Data and Methods} & $AP_{0.5}$ & $AP_{0.6}$ & $AP_{0.7}$ & $AP_{0.8}$ & $AP_{0.9}$ & $ mean AP$ & $AJI$ \\
			\hline
			\multirow{5}{*}{BBBC}&UNet-2& .5455 & .4645 & .4497 & .4212 & .2355 & .4233 & .5327 \\
			&UNet-3& .8863 &   .8197 & .7412 & \textbf{.5795} & \textbf{.2717} & \textbf{.6597} & \textbf{.7786} \\
			&StarDist & .3098 & .1410 & .0372 & .0011 & .000 & .0978 & .4499 \\
			&MRCNN & \textbf{.8953} & \textbf{.8629} & \textbf{.8111} & .5305 & .0382 & .6276 & .7580 \\
			&Ours&  \underline{\textbf{.9357}} &  \underline{\textbf{.9188}} & \underline{\textbf{.8648}} & \underline{\textbf{.7606}} & \underline{\textbf{.2904}} & \underline{\textbf{.7541}} & \underline{\textbf{.8442}}  \\
			\hline
			\multirow{5}{*}{OCC}&UNet-2& .1548 & .1303 & .1129 & .1055 & .1048 & .1217 & .2058 \\
			&UNet-3 & .7010 & .6071 & .5097 & .3767 & .1950 & .4779 & .5802 \\
			&StarDist& .6556 & .5566 & .4346 & .2970 & .1547 & .4197 &  .6927 \\
			&MRCNN& \underline{\textbf{.9277}} & \underline{\textbf{.9181}} & \underline{\textbf{.8870}} & \underline{\textbf{.8117}} & \underline{\textbf{.5564}} & \underline{\textbf{.8202}} & \underline{\textbf{.8412}} \\
			&Ours&  \textbf{.9230} &  \textbf{.8768} &  \textbf{.8007} &  \textbf{.6788} &  \textbf{.4349} &  \textbf{.7429} & \textbf{.8353} \\
			\hline
			\multirow{5}{*}{CCDB}&UNet-2 & .3698 & .3360 & .3049 & .2763 & .2228 & .3020 &.1185\\
			&UNet-3& .7307 & \textbf{.6774} & \textbf{.6210} & \underline{\textbf{.5153}} & \underline{\textbf{.2838}} & \textbf{.5656} &\textbf{.7148}\\
			&StarDist& \textbf{.7428} & .6532 & .4958 & .2685 & .0326 & .4386 & .6903\\
			&MRCNN& .6248 & .5691 & 4888 & 3476 & .0763 & .4213 & .5842\\
			&Ours& \underline{\textbf{.7968}} & \underline{\textbf{.7467}} & \underline{\textbf{.6767}} & \textbf{.4889} & \textbf{.2230} & \underline{\textbf{.5864}} & \underline{\textbf{.7601}}\\
			
			\hline
		\end{tabular}
	\end{table}
	
\begin{figure}
		\centering
		\includegraphics[width=.15\linewidth]{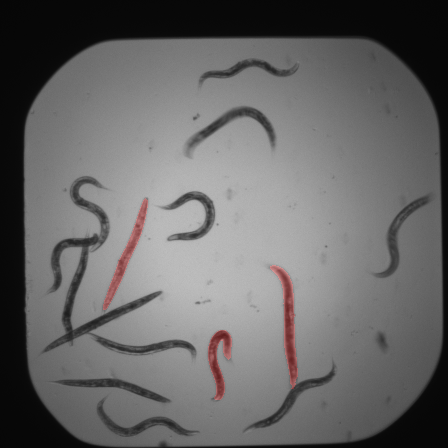}
		\includegraphics[width=.15\linewidth]{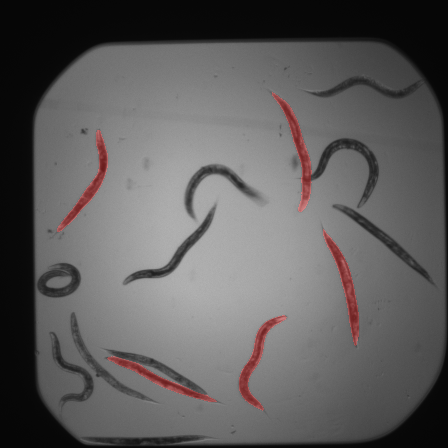}
		\includegraphics[width=.15\linewidth]{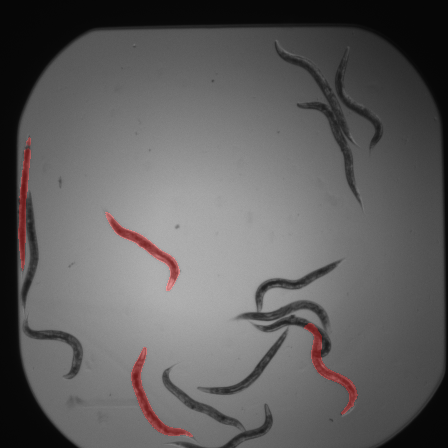}
		\includegraphics[width=.15\linewidth]{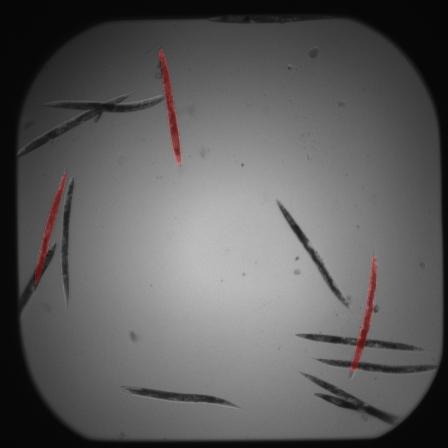}
		\includegraphics[width=.15\linewidth]{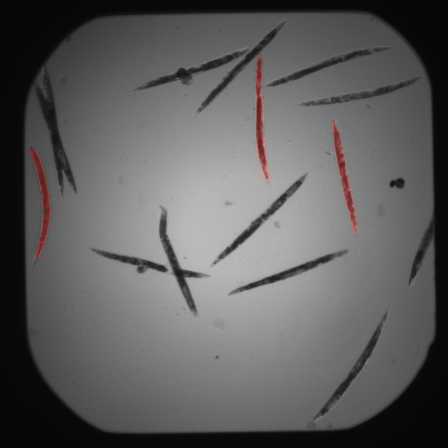}
		\includegraphics[width=.15\linewidth]{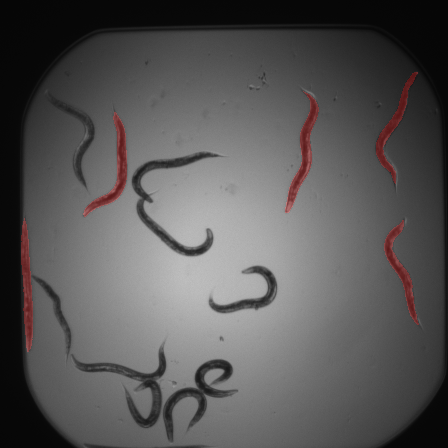}

		\includegraphics[width=.15\linewidth]{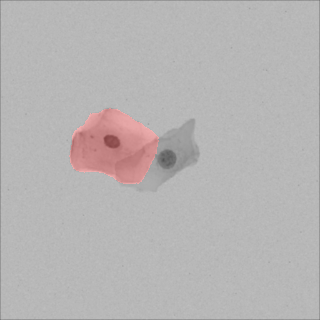}
		\includegraphics[width=.15\linewidth]{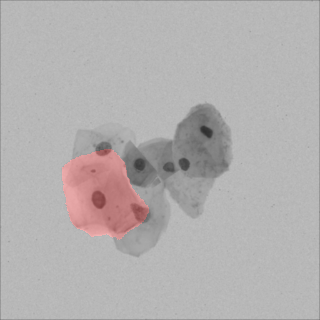}
		\includegraphics[width=.15\linewidth]{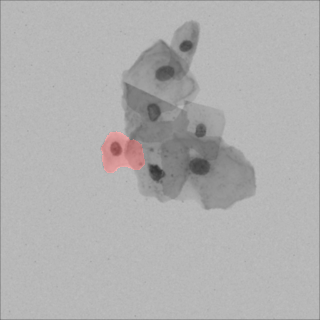}
		\includegraphics[width=.15\linewidth]{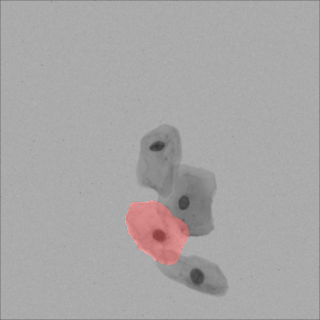}
		\includegraphics[width=.15\linewidth]{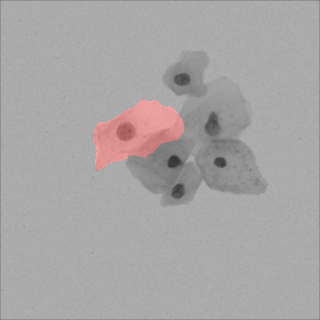}
		\includegraphics[width=.15\linewidth]{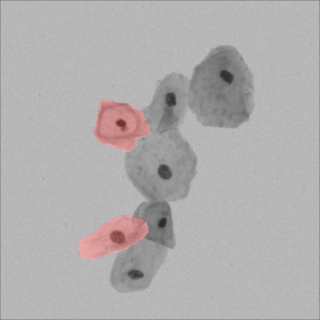}
		
		\caption{Example activation maps of certain layers. \emph{C. elegans} with the vertical body orientation are active in the 4th layer of the BBBC data (first row). In the 8th layer of OCC data (second row), most leftmost protruding cells are active. }
		\label{fig:channel}
	\end{figure}

	By chance, through some examples from experiments, we found certain layers reveal obvious semantics. For example, common morphological features, such as body orientation, are observed in layers predicted by the BBBC models (Figure~\ref{fig:channel}). Another feature commonly exploited by all models is the relative position. For example, the leftmost protruding objects are active in one common layer of OCC2014 predictions (Figure~\ref{fig:channel}). To capture high-level morphological features and sophisticated relative positions, especially in very crowded cases, an adequately large receptive field (RF) is required. To verify our analysis, we train three UNet variants with RF of 108, 220 and 444 pixels (Section~\ref{sec:implementation}). The experiments shows very significant differences: 39.49$\%$, 54.29$\%$ and 12.04$\%$ improvement from UNet-S to UNet-L on the datasets BBBC, OCC and CCDB in terms of $meanAP $ (Table~\ref{tab:study}).  
	
	In addition, we compared the performance before and after overlap completion (Table~\ref{tab:study}). On the dataset BBBC and CCDB, the difference is relatively small: 5.91$\%$ and 2.11$\%$, taking the best model UNet-L as an example, since the objects are only slightly overlapped. By contrast, the segmentation of severely overlapping cervical cells gains a 36.54$\%$ performance boost.
	
	\begin{table}
		\centering
		\caption{Performance of models with different sizes with/without overlap completion (OC) }
		\label{tab:study}
		\begin{tabular}{c|c|ccc}
			\hline
			\multicolumn{2}{c|}{$mean AP$} & UNet-S & UNet-M & UNet-L \\
			\hline
			\multirow{2}{*}{BBBC}& w/o OC & .4655 & .6670 & .7120  \\
			& w/ OC & .5406 & .7162 & \textbf{.7541} \\		
			\hline
			\multirow{2}{*}{OCC}& w/o OC & .4359 & .5273 & .5441  \\
			& w/ OC & .4815 & .6841 & \textbf{.7429} \\		
			\hline
			\multirow{2}{*}{CCDB}& w/o OC & .5229 & .5569 & .5743  \\
			& w/ OC & .5234 & .5614 & \textbf{.5864} \\		
			\hline
		\end{tabular}
	\end{table}
	
%
%
	
	\section{Conclusion and Outlook}
	\label{sec:conclusion}
	
	Our proposed approach can successfully layer touching and overlapping objects into different image layers. By grouping spatially separated objects in the same layer, our method simplifies post-processing and improves the accuracy of instance segmentation, yielding very competitive results on a diverse line of data sets. Our future work will focus on understanding the layering mechanism.

	\bibliography{egbib}

\begin{thebibliography}{24}
\providecommand{\natexlab}[1]{#1}
\providecommand{\url}[1]{\texttt{#1}}
\expandafter\ifx\csname urlstyle\endcsname\relax
  \providecommand{\doi}[1]{doi: #1}\else
  \providecommand{\doi}{doi: \begingroup \urlstyle{rm}\Url}\fi

\bibitem[Bai and Urtasun(2017)]{deepWatershed}
Min Bai and Raquel Urtasun.
\newblock Deep watershed transform for instance segmentation.
\newblock In \emph{Proceedings of the IEEE conference on computer vision and
  pattern recognition}, pages 5221--5229, 2017.

\bibitem[Chen et~al.(2016)Chen, Qi, Yu, and Heng]{dcan}
Hao Chen, Xiaojuan Qi, Lequan Yu, and Pheng-Ann Heng.
\newblock Dcan: deep contour-aware networks for accurate gland segmentation.
\newblock In \emph{Proceedings of the IEEE conference on Computer Vision and
  Pattern Recognition}, pages 2487--2496, 2016.

\bibitem[Chen et~al.(2019)Chen, Strauch, and Merhof]{cosEmb}
Long Chen, Martin Strauch, and Dorit Merhof.
\newblock Instance segmentation of biomedical images with an object-aware
  embedding learned with local constraints.
\newblock In \emph{International Conference on Medical Image Computing and
  Computer-Assisted Intervention}, pages 451--459. Springer, 2019.

\bibitem[Comaniciu and Meer(2002)]{meanShift}
Dorin Comaniciu and Peter Meer.
\newblock Mean shift: A robust approach toward feature space analysis.
\newblock \emph{IEEE Transactions on pattern analysis and machine
  intelligence}, 24\penalty0 (5):\penalty0 603--619, 2002.

\bibitem[De~Brabandere et~al.(2017)De~Brabandere, Neven, and Van~Gool]{disEmb}
Bert De~Brabandere, Davy Neven, and Luc Van~Gool.
\newblock Semantic instance segmentation for autonomous driving.
\newblock In \emph{Proceedings of the IEEE Conference on Computer Vision and
  Pattern Recognition Workshops}, pages 7--9, 2017.

\bibitem[Ester et~al.(1996)Ester, Kriegel, Sander, Xu, et~al.]{DBSCAN}
Martin Ester, Hans-Peter Kriegel, J{\"o}rg Sander, Xiaowei Xu, et~al.
\newblock A density-based algorithm for discovering clusters in large spatial
  databases with noise.
\newblock In \emph{Kdd}, volume~96, pages 226--231, 1996.

\bibitem[Gao et~al.(2019)Gao, Shan, Wang, Zhao, Yu, Yang, and Huang]{ssap}
Naiyu Gao, Yanhu Shan, Yupei Wang, Xin Zhao, Yinan Yu, Ming Yang, and Kaiqi
  Huang.
\newblock Ssap: Single-shot instance segmentation with affinity pyramid.
\newblock In \emph{Proceedings of the IEEE/CVF International Conference on
  Computer Vision}, pages 642--651, 2019.

\bibitem[Girshick et~al.(2014)Girshick, Donahue, Darrell, and Malik]{rcnn}
Ross Girshick, Jeff Donahue, Trevor Darrell, and Jitendra Malik.
\newblock Rich feature hierarchies for accurate object detection and semantic
  segmentation, 2014.

\bibitem[He et~al.(2016)He, Zhang, Ren, and Sun]{resnet}
Kaiming He, Xiangyu Zhang, Shaoqing Ren, and Jian Sun.
\newblock Deep residual learning for image recognition.
\newblock In \emph{Proceedings of the IEEE conference on computer vision and
  pattern recognition}, pages 770--778, 2016.

\bibitem[He et~al.(2018)He, Gkioxari, Dollár, and Girshick]{mrcnn}
Kaiming He, Georgia Gkioxari, Piotr Dollár, and Ross Girshick.
\newblock Mask r-cnn, 2018.

\bibitem[Hinton et~al.(2012)Hinton, Srivastava, and Swersky]{rmsprop}
Geoffrey Hinton, Nitish Srivastava, and Kevin Swersky.
\newblock Neural networks for machine learning lecture 6a overview of
  mini-batch gradient descent.
\newblock \emph{Cited on}, 14\penalty0 (8):\penalty0 2, 2012.

\bibitem[Kong and Fowlkes(2018)]{rnnEmb}
Shu Kong and Charless~C Fowlkes.
\newblock Recurrent pixel embedding for instance grouping.
\newblock In \emph{Proceedings of the IEEE Conference on Computer Vision and
  Pattern Recognition}, pages 9018--9028, 2018.

\bibitem[Kumar et~al.(2017)Kumar, Verma, Sharma, Bhargava, Vahadane, and
  Sethi]{aji}
Neeraj Kumar, Ruchika Verma, Sanuj Sharma, Surabhi Bhargava, Abhishek Vahadane,
  and Amit Sethi.
\newblock A dataset and a technique for generalized nuclear segmentation for
  computational pathology.
\newblock \emph{IEEE transactions on medical imaging}, 36\penalty0
  (7):\penalty0 1550--1560, 2017.

\bibitem[Lee et~al.(1994)Lee, Kashyap, and Chu]{skeleton}
Ta-Chih Lee, Rangasami~L. Kashyap, and Chong-Nam Chu.
\newblock Building skeleton models via 3-d medial surface/axis thinning
  algorithms.
\newblock \emph{Computer Vision, Graphics, and Image Processing}, 56\penalty0
  (6):\penalty0 462--478, 1994.

\bibitem[Lin et~al.(2014)Lin, Maire, Belongie, Hays, Perona, Ramanan,
  Doll{\'a}r, and Zitnick]{coco}
Tsung-Yi Lin, Michael Maire, Serge Belongie, James Hays, Pietro Perona, Deva
  Ramanan, Piotr Doll{\'a}r, and C~Lawrence Zitnick.
\newblock Microsoft coco: Common objects in context.
\newblock In \emph{European conference on computer vision}, pages 740--755.
  Springer, 2014.

\bibitem[Ljosa et~al.(2012)Ljosa, Sokolnicki, and Carpenter]{bbbc}
Vebjorn Ljosa, Katherine~L Sokolnicki, and Anne~E Carpenter.
\newblock Annotated high-throughput microscopy image sets for validation.
\newblock \emph{Nature methods}, 9\penalty0 (7):\penalty0 637--637, 2012.

\bibitem[Lu et~al.(2015)Lu, Carneiro, and Bradley]{occ14-B}
Zhi Lu, Gustavo Carneiro, and Andrew~P Bradley.
\newblock An improved joint optimization of multiple level set functions for
  the segmentation of overlapping cervical cells.
\newblock \emph{IEEE Transactions on Image Processing}, 24\penalty0
  (4):\penalty0 1261--1272, 2015.

\bibitem[Lu et~al.(2016)Lu, Carneiro, Bradley, Ushizima, Nosrati, Bianchi,
  Carneiro, and Hamarneh]{occ14-A}
Zhi Lu, Gustavo Carneiro, Andrew~P Bradley, Daniela Ushizima, Masoud~S Nosrati,
  Andrea~GC Bianchi, Claudia~M Carneiro, and Ghassan Hamarneh.
\newblock Evaluation of three algorithms for the segmentation of overlapping
  cervical cells.
\newblock \emph{IEEE journal of biomedical and health informatics}, 21\penalty0
  (2):\penalty0 441--450, 2016.

\bibitem[Milletari et~al.(2016)Milletari, Navab, and Ahmadi]{dice}
Fausto Milletari, Nassir Navab, and Seyed-Ahmad Ahmadi.
\newblock V-net: Fully convolutional neural networks for volumetric medical
  image segmentation.
\newblock In \emph{2016 fourth international conference on 3D vision (3DV)},
  pages 565--571. IEEE, 2016.

\bibitem[Ren et~al.(2015)Ren, He, Girshick, and Sun]{frcnn}
Shaoqing Ren, Kaiming He, Ross Girshick, and Jian Sun.
\newblock Faster r-cnn: Towards real-time object detection with region proposal
  networks.
\newblock \emph{Advances in neural information processing systems}, 28, 2015.

\bibitem[Ronneberger et~al.(2015)Ronneberger, Fischer, and Brox]{unet}
Olaf Ronneberger, Philipp Fischer, and Thomas Brox.
\newblock U-net: Convolutional networks for biomedical image segmentation.
\newblock In \emph{International Conference on Medical image computing and
  computer-assisted intervention}, pages 234--241. Springer, 2015.

\bibitem[Schmidt et~al.(2018)Schmidt, Weigert, Broaddus, and Myers]{StarDist}
Uwe Schmidt, Martin Weigert, Coleman Broaddus, and Gene Myers.
\newblock Cell detection with star-convex polygons.
\newblock In \emph{International Conference on Medical Image Computing and
  Computer-Assisted Intervention}, pages 265--273. Springer, 2018.

\bibitem[Wu et~al.(2020)Wu, Chen, and Merhof]{impemb}
Yuli Wu, Long Chen, and Dorit Merhof.
\newblock Improving pixel embedding learning through intermediate distance
  regression supervision for instance segmentation.
\newblock In \emph{ECCV Workshops}, pages 213--227, 2020.

\bibitem[Yu et~al.()Yu, Lee, Hariharan, Bu, and Ahmed]{ccdb}
Weimiao Yu, Hwee~Kuan Lee, Srivats Hariharan, Wen~Yu Bu, and Sohail Ahmed.
\newblock Ccdb: 6843, mus musculus, neuroblastoma.
\newblock \emph{Cell Image Library.}
\newblock URL \url{https://doi.org/doi:10.7295/W9CCDB6843}.

\end{thebibliography}
\end{document}